\documentclass[letterpaper, 10 pt, conference]{ieeeconf}  % Comment this line out if you need a4paper

\IEEEoverridecommandlockouts                              % This command is only needed if 
\usepackage[utf8]{inputenc}
\usepackage{graphics} % for pdf, bitmapped graphics files
\usepackage{epsfig} % for postscript graphics files
\usepackage{amsmath,amssymb,amsfonts}
\usepackage{mathtools}
\usepackage{algorithm}
\usepackage{multirow}
\usepackage{booktabs}
\usepackage{algpseudocode}
\usepackage{tikz}
\usetikzlibrary{arrows.meta, positioning}
\usepackage{bm}
\usepackage{subcaption}
\usepackage{colortbl}
\usepackage{makecell}
\usepackage[nospace,compress, sort]{cite}

\usepackage{hyperref}
\usepackage{siunitx}
\usepackage{graphicx}
\usepackage{subcaption}
\usepackage{afterpage}
\usepackage{cancel}
\usepackage{threeparttable}
\usepackage{xcolor}

\newif\ifshowchangesone
\showchangesonefalse
\newif\ifshowchangestwo
\showchangestwofalse

\newif\ifshowchangesthree
\showchangesthreefalse

\newcommand{\revone}[1]{%
  \ifshowchangesone
    \textcolor{blue}{#1}%
  \else
    #1%
  \fi
}

\newcommand{\revtwo}[1]{%
  \ifshowchangestwo
    \textcolor{blue}{#1}%
  \else
    #1%
  \fi
}

\newcommand{\revthree}[1]{%
  \ifshowchangesthree
    \textcolor{blue}{#1}%
  \else
    #1%
  \fi
}
\usepackage{pgfplots}
\DeclareUnicodeCharacter{2212}{−}
\usepgfplotslibrary{groupplots,dateplot}
\usetikzlibrary{patterns,shapes.arrows}
\pgfplotsset{compat=newest}

\DeclareMathAlphabet{\mathcal}{OMS}{cmsy}{m}{n}

\renewcommand{\vec}{\bm}

\newcommand{\R}{\mathbb{R}}

\title{\LARGE \bf
SQ-CBF: Signed Distance Functions for \revtwo{Reliable} \\Superquadric-Based Safety Filtering
}

\author{
Haocheng Zhao$^{*}$,
Lukas Brunke$^{*}$,
Oliver Lagerquist,
Siqi Zhou,
and Angela P. Schoellig
\thanks{$^{*}$Equal contribution.}
\thanks{Haocheng Zhao is with the Learning Systems and Robotics Lab, Technical University of Munich, 80333 Munich, Germany. Email: haocheng.zhao@tum.de}%
\thanks{Lukas Brunke and Angela P. Schoellig are with the Learning Systems and Robotics Lab, Technical University of Munich, 80333 Munich, Germany, also with the University of Toronto Institute for Aerospace Studies, North York, ON M3H 5T6, Canada, and also with the Vector Institute for Artificial Intelligence, Toronto, ON M5G 0C6, Canada. Emails: \{lukas.brunke, angela.schoellig\}@tum.de}%
\thanks{Oliver Lagerquist is with the Learning Systems and Robotics Lab, Technical University of Munich, 80333 Munich, Germany, and also with the University of Toronto Institute for Aerospace Studies, North York, ON M3H 5T6, Canada. Email: oliver.lagerquist@mail.utoronto.ca}%
\thanks{Siqi Zhou is with the Learning Systems and Robotics Lab, Technical University of Munich, 80333 Munich, Germany, and also with Simon Fraser University, Burnaby, BC V5A 1S6, Canada. Email: siqi@sfu.ca}
}

\begin{document}

\maketitle
\thispagestyle{empty}
\pagestyle{empty}

%%%%%%%%%%%%%%%%%%%%%%%%%%%%%%%%%%%%%%%%%%%%%%%%%%%%%%%%%%%%%%%%%%%%%%%%%%%%%%%%

\begin{abstract}
Ensuring safe robot operation in cluttered and dynamic environments remains a fundamental challenge. While control barrier functions (CBFs) provide an effective framework for real-time safety filtering, their performance critically depends on the underlying geometric representation, which is often simplified, leading to either overly conservative behavior or insufficient collision coverage. Superquadrics~(SQs) offer an expressive way to model complex shapes using a few primitives and are increasingly used for robot safety. To integrate this representation into CBF, most existing approaches directly use their implicit functions as barrier candidates. However, we identify a critical but overlooked issue in this practice: the gradients of the implicit SQ function can become severely ill-conditioned, potentially rendering the optimization infeasible and undermining reliable safety filtering. To address this issue, we formulate an SQ-based safety filter that uses signed distance functions (SDFs) as barrier candidates. Since analytical SDFs are unavailable for general SQs, we compute the distance using the efficient Gilbert-Johnson-Keerthi algorithm and \revone{derive its gradient in closed form}. Extensive simulation and real-world experiments demonstrate consistent collision-free manipulation in cluttered scenes, showing robustness to challenging geometries, sensing noise, and dynamic disturbances, while improving task efficiency in teleoperation tasks. These results highlight a pathway toward safety filters that remain precise and reliable under the geometric complexity of real-world environments.
\end{abstract}

%%%%%%%%%%%%%%%%%%%%%%%%%%%%%%%%%%%%%%%%%%%%%%%%%%%%%%%%%%%%%%%%%%%%%%%%%%%%%%%%
\section{Introduction}

A central challenge in robotics is transferring manipulators from structured laboratory settings to cluttered, dynamic real-world scenes. To ensure safe yet efficient operation alongside humans and obstacles, control algorithms must prevent unsafe behaviors without compromising task performance. In this context, control barrier functions~(CBFs)~\cite{Ames2019CBFReview} have matured into a standard tool for safety filtering~\cite{Brunke2022SafeLearning}, as they can enforce safety guarantees with low computational overhead. A safety filter takes an unverified control input~(e.g., teleoperation or a motion policy) and minimizes modifications to it while ensuring safety.
However, its performance depends critically on the geometric representation.
Ensuring safe operation in complex environments often necessitates a trade-off between geometric fidelity and computational efficiency.
Simple primitives such as spheres are efficient and analytically tractable but overly conservative, limiting operation in tight spaces~\cite{Zimmermann2022DifferentiableCA}.
Conversely, high-fidelity representations such as meshes~\cite{Singletary2022RAL} and Gaussian splats~\cite{Chen2025ICRA} capture geometric detail but incur significant computational cost and may yield uninformative gradients in non-convex regions, leading to oscillatory or unstable behavior.

To bridge this gap, we require a representation that is both compact and expressive. 
Superquadrics (SQs)~\cite{Barr1981Superquadrics} provide an alternative, unifying common primitives while maintaining differentiability. 
Recent advances in perception further enable efficient SQ fitting from sensor data~\cite{Liu2022SuperquadricRecovery,fedele2025superdec}.
However, prior work often employs SQ implicit functions as CBF candidates~\cite{Xu2018ICRA,Xiao2023TRO,Lukas2025SemanticallySafety}, which are not true distance metrics and can exhibit unbounded gradients as geometries become sharp~(e.g., box-like) due to the exponential terms.
This leads to poor numerical conditioning of the underlying optimization~\cite{Narunas2019TPAMI} and can render the safety filter infeasible.

\begin{figure}[t]
    \centering
        \includegraphics[
            width=\linewidth,
            trim=5 5 5 5,   % left bottom right top
            clip
        ]{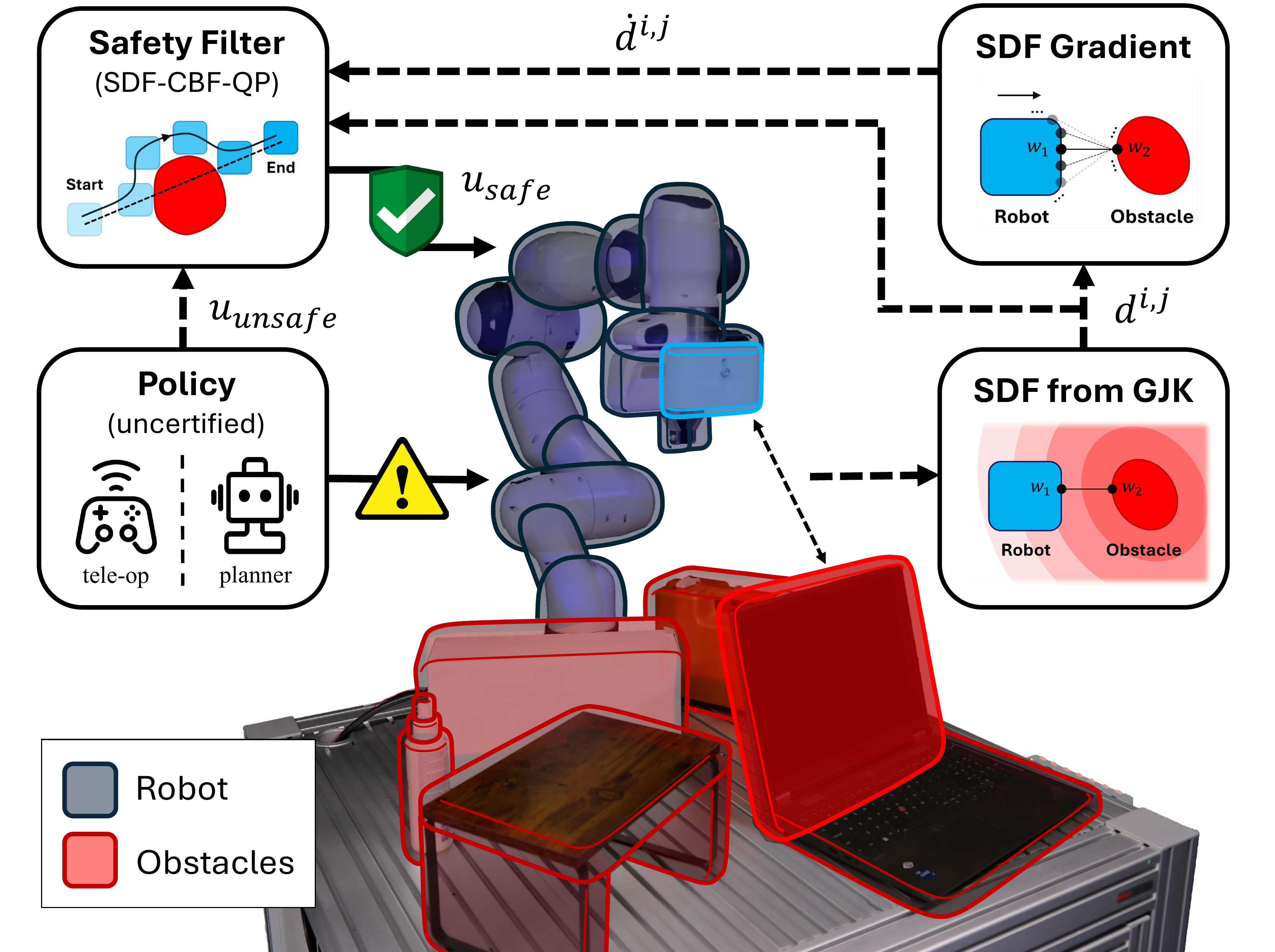}
        \caption{Overview of the proposed SQ-CBF safety filter. The framework represents both the robot (blue) and the obstacles (red) using compact superquadrics and enforces collision avoidance through an SDF-based CBF. SDFs are computed from superquadric geometries using GJK/EPA, while well-behaved distance gradients are estimated online and incorporated into the safety filter. Given an unverified control command~(e.g., from teleoperation or a planner), the safety filter modifies the command online by solving a CBF-QP, intervening only when safety is at risk. A video demonstrating the safety filter's performance can be found here: \href{http://tiny.cc/sq-cbf}{http://tiny.cc/sq-cbf}.}
    \label{fig:money_fig}
\end{figure}

In this work, we focus on SQ-based geometric representations for CBF safety filtering in the context of collision avoidance. We adopt the signed distance function (SDF) as the CBF candidate,
which maintains a gradient norm of~1 almost everywhere. %~\cite{OsherFedkiw2003}. 
\revone{Since evaluating the SDF between true SQs requires solving a nonlinear constrained optimization problem and is computationally expensive, we 
use the Gilbert-Johnson-Keerthi~(GJK) algorithm~\cite{GJK1988} for efficient SDF evaluation.}
This avoids the poor conditioning of implicit-function-based barriers and improves the reliability of safety filtering.
\revtwo{With the full pipeline illustrated in~\autoref{fig:money_fig}, we summarize our contributions as follows:
\begin{enumerate}
    \item We identify a long-overlooked issue in applying SQs to gradient-based collision avoidance: while efficient, its implicit functions yield poorly conditioned gradients, which lead to infeasibility of the optimization problem. 
    \item To address this issue, we formulate an SQ-based CBF framework that uses the SDF between SQs as the barrier function, and derive a closed-form expression for its derivative with respect to the SQ poses.
    \item We develop a practical implementation that evaluates the SDF between SQs in real time through dense polytopic approximations and GJK/EPA, together with a robustness analysis that explicitly accounts for the approximation-induced errors.
    \item We validate the proposed safety filter through extensive sim- and real-experiments, demonstrating its efficacy in cluttered, unstructured, and dynamic environments.
\end{enumerate}
}

\section{Related Work}
\label{sec:related-works}

\subsection{Geometric Representations for Safe Robot Control}

Faithful geometric representation of the robot and environment inherently involves a trade-off between geometric fidelity and computational efficiency.
A common strategy is to approximate the geometries using simple primitives such as spheres~\cite{Daniel2025OSCBF}, capsules~\cite{Zimmermann2022DifferentiableCA}, cylinders or boxes~\cite{Vinicius2024SmoothDistances}, and ellipsoids~\cite{Dai2023RAL}.
Primitive-based representations trade off expressiveness for efficiency, often leading to
volume over-approximation (conservative)~\cite{Zimmermann2022DifferentiableCA}, under-coverage (unsafe)~\cite{Daniel2025OSCBF}, or poor scalability in optimization~\cite{Chen2025ICRA}. 
Conversely, high-fidelity models enable more accurate geometric reasoning and collision checking, and can be obtained from CAD meshes~\cite{Singletary2022RAL}, point clouds~\cite{DeSa2024ICRA}, volumetric maps~\cite{Zhou2024ICRA}, and Gaussian splatting~\cite{Chen2025ICRA}.
However, incorporating these representations in real-time CBF-based safety filtering often relies on additional processing~(e.g., convex decomposition, translating into primitives~\cite{Lukas2025SemanticallySafety}, or pruning~\cite{Chen2025ICRA}) to enable efficient distance queries and well-behaved gradients.

\subsection{Superquadrics in Robotics}
\label{subsec:SQinRob}

In this work, we utilize SQs~\cite{Barr1981Superquadrics} for collision checking.
Their high geometric expressiveness enables accurate representation of complex geometries with only a small number of primitives.
SQs have been widely used in robotics for applications including 
aerial navigation~\cite{Xu2018ICRA}, autonomous driving~\cite{Xiao2023TRO, Lu2025SeparatingHyperplane}, and safe semantic manipulation~\cite{Lukas2025SemanticallySafety}.
Meanwhile, advances in computer vision now enable SQs to be directly recovered from point clouds~\cite{Liu2022SuperquadricRecovery,fedele2025superdec} and RGB images~\cite{gao2025selfsupervisedlearninghybridpartaware}, facilitating their broader adoption in robotics.
However, most prior CBF works rely on the SQ implicit function~\cite{Xu2018ICRA,Xiao2023TRO,Lukas2025SemanticallySafety},
which is not a distance metric, and its gradient norm may become unbounded for large separations or for shape parameters approaching sharp geometries~\cite{Narunas2019TPAMI}.
\revtwo{This numerical conditioning issue} makes the implicit function unsuitable for an optimization-based safety filter, as it can render the underlying problem infeasible.
Instead, we formulate the CBF using the SDF $d(\vec{x})$, which
satisfies the eikonal equation with $\|\nabla d(\vec{x})\| = 1$ almost everywhere~\cite{Singletary2022RAL}, yielding bounded and geometrically meaningful gradients.

\subsection{Distance Functions and Differentiability}

For simple geometries such as spheres, the SDF can be computed analytically with negligible cost.
\revone{
However, for complex geometries such as SQs, SDF evaluation generally lacks a closed-form solution and requires iterative constrained optimization, making it computationally expensive for online safety filtering.
For general ellipsoids, separating-hyperplane-based CBF formulations can avoid direct SDF evaluation~\cite{Lu2025SeparatingHyperplane}, but they still rely on the closed-form ellipsoidal representation and are not directly applicable to SQs.
Prior work also leverages the rigid body radial Euclidean distance as an efficient proxy~\cite{Louis2025CASE}, but it may miss collisions when closest points do not lie along the center-connecting direction.}
Alternatively, for convex polytopes, SDF values are typically returned by the GJK algorithm and the expanding polytope algorithm~(EPA)~\cite{GJK1988, VanDenBergen2004EPA}.
However, differentiating through GJK is still challenging due to discontinuities at vertex/face transitions. 
\revone{In our approach, we discretize the SQ surfaces into polytopic approximations, enabling GJK-based efficient distance evaluation, and derive the closed-form SDF gradient with respect to the configuration.}
This enables an SQ-based safety filter that is more efficient yet \revtwo{reliable}.

% END

\section{Problem Definition}

We consider the collision avoidance for a robot in a potentially changing scene, where both the robot and obstacles are represented by geometric models.
The robot is modeled as a velocity-controlled manipulator with joint configuration $\vec{q} \in \mathbb{R}^n$ and kinematic model $\dot{\vec{q}}=\vec{u}$, where $\vec{u} \in \mathbb{R}^n$ is the joint velocity command and forward kinematics uniquely determine the 6D poses of all attached collision geometries.
Each obstacle geometry is described by its 6D pose \revone{$\vec{\xi}_{\mathrm{obs}}=\big[ \vec{t}^\top_{\mathrm{obs}},  \vec{\theta}^\top_{\mathrm{obs}} \big]^\top\in \mathbb{R}^{6}$} expressed in world frame $\mathcal{W}$. These poses may change over time according to updates from perception.
\revone{The full system state is defined as
$\vec{x}^{\top} = \big[ \vec{q}^{\top}, \mathbf{\Xi}_{\mathrm{obs}}^{\top}, \dot{\mathbf{\Xi}}_{\mathrm{obs}}^{\top} \big]$, where 
$\mathbf{\Xi}_{\mathrm{obs}}^{\top} = \big[ (\vec{\xi}_{\mathrm{obs}}^{1})^{\top}, \cdots, (\vec{\xi}_{\mathrm{obs}}^{q})^{\top} \big]$
stacks the poses of all obstacle geometries. In static scenes, constant obstacle states are omitted and $\vec{x}=\vec{q}$.}
Let $d^{i,j}(\vec{q}, \vec{\xi}^j_{\mathrm{obs}})$ denote the signed distance between the $i$-th robot collision geometry and the $j$-th obstacle geometry,
the safe set ensuring collision avoidance can be constructed as $\mathcal{S} = \{\vec{x} \;|\; d^{i,j} \ge 0, \;\forall i,j\}.$
\revtwo{Throughout this paper, we assess the reliability of the safety filter framework primarily in terms of the numerical conditioning of the quantities (i.e., barrier values and gradients) entering the CBF conditions.}

\section{Preliminaries}
\label{sec:background}

\subsection{Superquadrics}

Primitives such as ellipsoids, cylinders, and boxes can be generalized via SQs using only a small set of scale and shape parameters~\cite{Barr1981Superquadrics}. 
An axis-aligned SQ centered at the origin is defined by the implicit function
\revone{
\begin{equation}
  f_{\mathrm{sq}}(\vec{p}; \vec{s}) =
  \left[
    \left|\frac{x}{a_1}\right|^{\frac{2}{e_2}} +
    \left|\frac{y}{a_2}\right|^{\frac{2}{e_2}}
  \right]^{\frac{e_2}{e_1}} +
  \left|\frac{z}{a_3}\right|^{\frac{2}{e_1}} - 1,
  \label{eq:sq_implicit}
\end{equation}}%
where $\vec{p} = [x, y, z]^\top$ is the Cartesian coordinates and
$\vec{s} = [a_1, a_2, a_3, e_1, e_2]^\top$ is the shape parameters.
\revone{Absolute value operators $|\cdot|$ are used to ensure that the function remains real-valued for all $\vec{p} \in \mathbb{R}^3$ when fractional exponents are applied.}
SQs are strictly convex for $e_{1,2} \in (0,2)$ and become concave otherwise~\cite{Seoki2024SQC2continuity}. 
We represent the set of points on and within the SQ as the 0-sublevel set 
\revone{$\mathcal{Q} = \{\vec{p} : f_{\mathrm{sq}}(\vec{p}; \vec{s}) \le 0\}$}.

\begin{figure}
    \centering
    \begin{subfigure}[t]{0.48\textwidth}
        \centering
        \input{tikz/unbounded_function_and_gradient}
    \end{subfigure}
    \hfill
    \begin{subfigure}[t]{0.45\textwidth}
        \hspace*{-0.01\textwidth} % offset to right side
        \includegraphics[width=\linewidth]{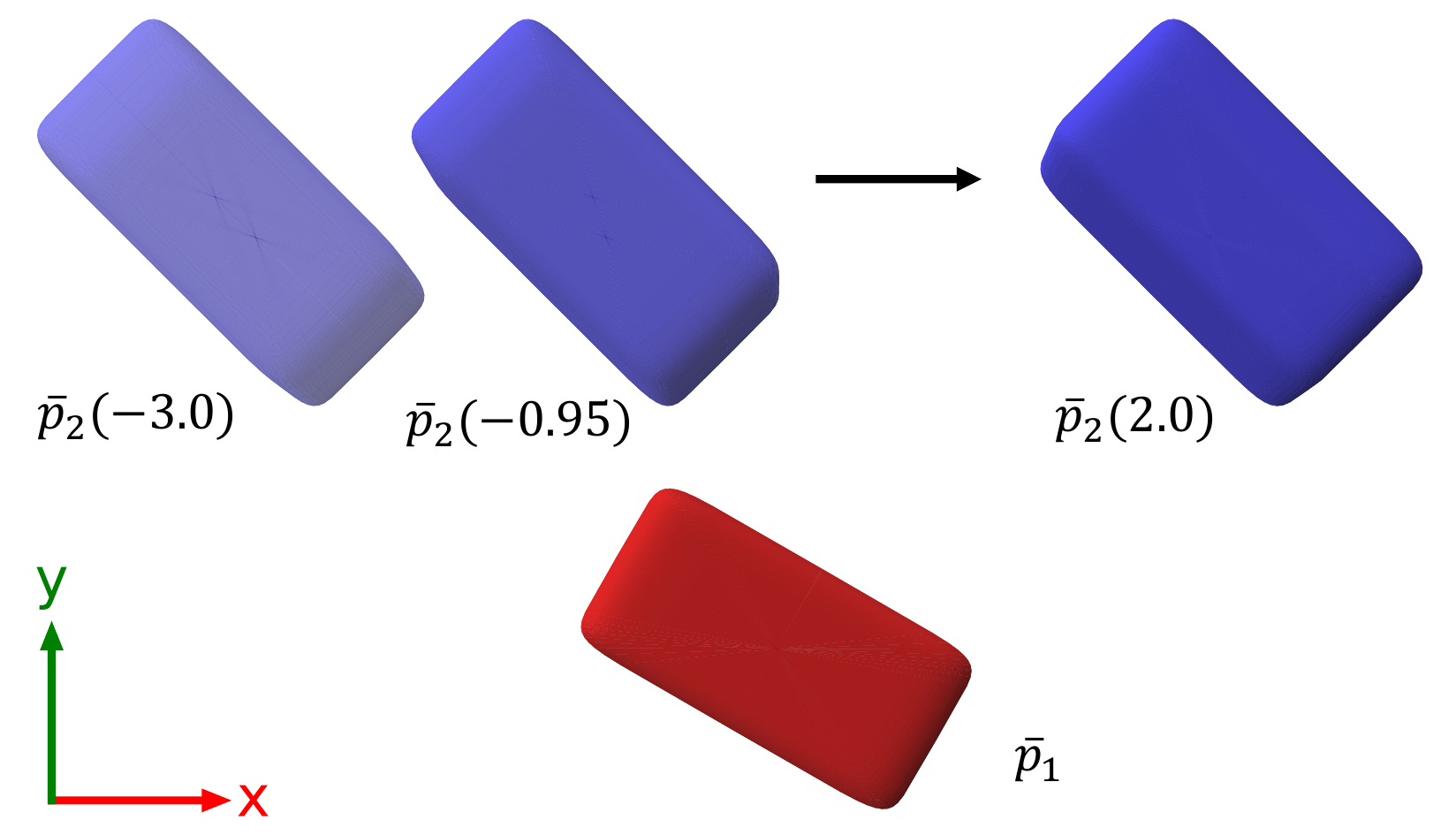}
    \end{subfigure}
    \caption{Evaluating the implicit function $f_{\text{sq}}^*(x) = \min_{\vec{p} \in \R^3}$ $f_{\text{sq}}(\vec{R}_z(\pi / 3)^\top (\vec{p} - \bar{\vec{p}}_1), \vec{s}_1) ~ \text{s.t.} ~f_{\text{sq}}(\vec{R}_z(-\pi / 4)^\top(\vec{p} - \bar{\vec{p}}_2(x)), \vec{s}_2) \leq 1$ from~\cite{Wei2026TCST} 
    % ($\vec{R}_z(\varphi)$ is a rotation around the $z$-axis by $\varphi$) 
    and the SDF $d(x)$ and their gradients for two superquadrics with parameters $\vec{s
    }_1 = [0.5, 1.5, 1.0, 0.2, 0.2]^\top$ and $\vec{s
    }_2 = [1.0, 0.5, 1.0, 0.2, 0.2]^\top$ while varying their relative $x$ position with $\bar{\vec{p}}_1 - \bar{\vec{p}}_2(x) = [0, 0, 0]^\top - [x, 3, 0]^\top$. The implicit function and its gradient are large~(greater than $10^4$), leading to ill-conditioned matrices in the CBF-QP. In practice, this may render the optimization infeasible, which is further demonstrated by the open-loop comparison presented in~\autoref{fig:feasibility_comparison}.}
    \label{fig:inside-outside}
\end{figure}

As discussed in \autoref{subsec:SQinRob}, many prior works leverage SQs to enforce safety constraints as they offer an expressive representation with continuously differentiable property
(assuming
$e_{1,2} \in \left(0, 2\right)$).
However, directly using the implicit function~\eqref{eq:sq_implicit} to distinguish between interior and exterior regions is undesirable in gradient-based optimization, as 
$f_{\text{sq}}$ and its gradient may grow unboundedly~(see~\autoref{fig:inside-outside}).
These are caused by small exponential coefficients $(e_1, e_2)$
or when evaluating points $\vec{p}$ far from the SQ's boundary $\partial\mathcal{Q}$~\cite{Narunas2019TPAMI}.

\subsection{CBF-QP Formulation}

% CBF
To formally guarantee safety in continuous control systems, CBFs provide a framework that ensures the system initialized within a desired set remains within the set for all future time, i.e., the set is positively forward invariant
~\cite{Ames2019CBFReview}.
Let $\mathcal{S}$ denote a safe set defined by the 0-superlevel set of a continuously differentiable function $h:\mathbb{R}^n \to \mathbb{R}$, i.e.,
$\mathcal{S} = \{ \vec{x} \in \mathbb{R}^n \mid h(\vec{x}) \ge 0 \}.$
To ensure that all trajectories starting within $\mathcal{S}$ remain inside for all future time, the system is required to satisfy the CBF condition:
  $\sup_{\vec{u} \in \mathcal{U}} \dot{h}(\vec{x}, \vec{u}) \ge - \alpha(h(\vec{x}))$, 
\noindent where 
$\vec{u} \in \mathcal{U}$ denotes the admissible control input, $\dot{h}(\vec{x}, \vec{u})$
denotes the time derivative of $h(\vec{x})$, and $\alpha(\cdot)$ is an extended class-$\mathcal{K}$ function. 

In the context of safety-critical control, the safe control input can be obtained by solving the following optimization problem that enforces the CBF constraint:
\begin{align}
  \vec{u}^*(\vec{x}, \vec{u}_{\mathrm{cmd}}) = \; \arg & \min_{\vec{u} \in \mathcal{U}} \quad 
  \|\vec{u} - \vec{u}_{\mathrm{cmd}}\|^2_2  \label{eq:cbf_qp}\\
  \text{s.t.} & \; \dot{h}(\vec{x}, \vec{u}) \ge - \alpha(h(\vec{x})), \nonumber
\end{align}

\noindent where $\vec{u}_{\mathrm{cmd}}$ is a nominal input.
\revone{For control-affine systems $\dot{\vec{x}} = f(\vec{x}) + g(\vec{x}) \vec{u}$, the time derivative $\dot{h}(\vec{x}, \vec{u}) = L_f h(\vec{x}) + L_g h(\vec{x}) \vec{u}$ is also affine in $\vec{u}$, and problem~\eqref{eq:cbf_qp} reduces to a quadratic program (QP), referred to as a CBF-QP~\cite{Ames2019CBFReview}.}
The resulting CBF-QP keeps the system's state within $\mathcal{S}$ while minimally altering the intended behavior.

\section{Methodology}

\subsection{Robot and Environment  Representation}

Both the robot and the environment are represented by SQs, which provide a compact yet expressive approximation of complex scenes. 
The robot is modeled as a collection of SQs rigidly attached to its kinematic chain~(see~\autoref{tab:case_comparison}), denoted by $\mathcal{Q}_{\mathrm{rob}}^i(\vec{q})$ for $i \in \mathcal{I}^{\mathrm{rob}} \triangleq \{1,\dots,p\}$. 
The $i$-th robot SQ is attached to link $\ell(i) \in \{1,\dots,n\}$ with fixed position and orientation offsets $\vec{p}_{\ell(i)\!\to i}$ and $\vec{\phi}_{\ell(i)\!\to i}$, respectively, relative to the link frame.
As a result, the 6D pose of each robot SQ is uniquely determined by the robot's forward kinematics. Meanwhile, obstacles in the environment are modeled as SQs with free 6D poses \revone{$\mathcal{Q}_{\mathrm{obs}}^j(\vec{\xi}^j_{\mathrm{obs}})$} for $j \in \mathcal{I}^{\mathrm{obs}} \triangleq \{p+1,\dots,p+q\}$.
In practice, such obstacle SQs can be fitted based on the RGB-D point clouds from perception.
Note that in this work, all SQs are restricted to strictly convex parameter regime (i.e., $e_{1,2} \in (0,2)$), ensuring valid and unique signed distance evaluation.

\subsection{Signed Distance Function as CBF}
\label{subsec:SdfCbf}

Consider two SQs $\mathcal{Q}^i$ and 
$\mathcal{Q}^j$ with $i \neq j$,
the SDF between these two convex shapes can be formulated as a 
constrained minimization problem
$d^{i,j} = s^{i,j} \min_{\Delta \vec{p} \in \partial \mathcal{M}^{i,j}} \|\Delta \vec{p}\|$, 
where 
$\mathcal{M}^{i, j}
= \{\, \Delta\vec{p} = \vec{p}_i - \vec{p}_j \mid \vec{p}_i \in \mathcal{Q}^i, \vec{p}_j \in \mathcal{Q}^j \,\}$ is the Minkowski difference and $s^{i,j} = 1$ if $\mathbf{0} \notin \mathcal{M}^{i,j}$ and $s^{i,j} = -1$ if $\mathbf{0} \in \mathcal{M}^{i,j}$.
Geometrically, $d^{i,j}>0$ indicates separation between the two shapes, 
$d^{i,j}=0$ corresponds to tangential contact, and $d^{i,j}<0$ denotes interpenetration.
A CBF candidate based on $d^{i,j}$ can be formulated as 
$h^{i,j}_{\delta}
= d^{i,j} - \epsilon_{\delta}$,
where $\epsilon_{\delta} > 0$ denotes a safety margin.
Enforcing \(h_{\delta}^{i,j}\geq0\) guarantees a strictly positive separation margin and thus collision-free motion.
\revone{By Theorem 3.5 in~\cite{Gilbert1985C1}, the Euclidean distance between two strictly convex bodies is continuously differentiable on positively separated configurations. 
Since SQs with $e_{1,2}\in(0,2)$ are strictly convex~\cite{Seoki2024SQC2continuity}, the SDF between $\mathcal{Q}^i$ and $\mathcal{Q}^j$ is continuously differentiable on $h_{\delta}^{i,j}\geq0$, making $h_{\delta}^{i,j}$ a valid CBF candidate on the safe set.}

\revone{
\subsection{Gradient of the Signed Distance Function}
\label{subsec:SdfGradEst}

We exploit the geometry of support functions and separation vectors to derive the SDF gradient w.r.t. the SQ's configuration in closed form. 
Consider two convex SQs $\mathcal{Q}^i$ and $\mathcal{Q}^j$ defined in local frames $\mathcal{F}^i$ and $\mathcal{F}^j$. 
Let $\sigma_{\mathcal{Q}}(\cdot)$ denote the support function and $\nabla\sigma_{\mathcal{Q}}(\cdot)$ its gradient, which returns a support point on $\partial\mathcal{Q}$. 
\revthree{
For SQs with $e_{1,2}\in(1,2)$, the support function is globally $\mathcal{C}^2$ continuous, whereas for SQs with $e_{1,2}\in(0,1]$, its Hessian may be undefined at certain support directions, e.g., six axis directions $(\pm a_1,0,0)$, $(0,\pm a_2,0)$, and $(0,0,\pm a_3)$~\cite{Seoki2024SQC2continuity}.
However, as shown at the end of this subsection, the final SDF gradient expression remains valid at these degenerate configurations, which therefore do not introduce singularities into the proposed method.
}
Each SQ is parameterized by $\boldsymbol{\xi}=[\vec{t}^\top,\vec{\theta}^\top]^\top$, with rotation matrix $\vec{R}\in SO(3)$ associated with $\vec{\theta}$. 
For notational simplicity, we focus on pair $(i,j)$ and omit superscripts in $\vec{J}^{i,j}_{\vec{\xi}^j}$ and $d^{i,j}$.

\paragraph{Derivation of the SDF gradient}
The optimal separation vector $\Delta \vec{p}^*\in\mathbb{R}^3$ expressed in $\mathcal{W}$ is governed by the stationarity condition 
$
\vec{f}_s(\Delta \vec{p}^*,\vec{\xi}^i,\vec{\xi}^j)
=
\Delta \vec{p}^*
-(\vec{t}_i+\vec{R}_i\,\vec{p}^*_i)
+(\vec{t}_j+\vec{R}_j\,\vec{p}^*_j) = \vec{0},
$
where $\vec{p}^*_i = \nabla \sigma_{\mathcal{Q}^i}(- \vec{R}_i^\top \Delta \vec{p}^*)$ and $\vec{p}^*_j = \nabla \sigma_{\mathcal{Q}^j}(\vec{R}_j^\top \Delta \vec{p}^*)$ denote the witness points
expressed in the local frames.
The gradient of $ \Delta\vec{p}^*$ w.r.t. $\vec{\xi}^j$ can be derived using the implicit function theorem~\cite{Montaut2023RandomizedSmoothing}:
\begin{equation}
\frac{\partial \Delta \vec{p}^*}{\partial \vec{\xi}^j}
=
-\Big(\frac{\partial \vec{f}_s}{\partial \Delta \vec{p}^*}\Big)^{-1}
\left.\frac{\partial \vec{f}_s}{\partial \vec{\xi}^j}\right|_{\Delta \vec{p}^*}.
\label{eq:implicit_cond}
\end{equation}
The Jacobian $\partial \vec{f}_s/\partial \Delta \vec{p}^*$ admits the closed-form:
\begin{equation}
\frac{\partial \vec{f}_s}{\partial \Delta \vec{p}^*}
=
\vec{I}
+
\vec{R}_i \vec{H}_i \vec{R}_i^\top
+
\vec{R}_j \vec{H}_j \vec{R}_j^\top,
\label{eq:df_ddp}
\end{equation}
where $\vec{H}_i := \nabla^2 \sigma_{\mathcal{Q}^i}(- \vec{R}_i^\top \Delta \vec{p}^*)$ and $\vec{H}_j:= \nabla^2 \sigma_{\mathcal{Q}^j}(\vec{R}_j^\top \Delta \vec{p}^*)$ denote the Hessians of the support functions.
By the chain rule, we obtain
\begin{equation}
\frac{\partial \vec{f}_s}{\partial \vec{\xi}^j}
=
\big[\,
\vec{I} \quad 
- \vec{R}_j\big( 
\,[\vec{p}_j^*]_\times + \vec{H}_j [\vec{R}_j^\top \Delta\vec{p}^*]_\times\,
\big) \vec{J}_r(\vec{\theta}_j)
\,\big],
\label{eq:df_dx}
\end{equation}
where $[\cdot]_{\times}$ denotes the hat operator in screw theory, and $\vec{J}_r(\cdot)$ is the right Jacobian following the convention in~\cite{Joan2018MircoLie}.
Based on $\Delta \vec{p}^*$, the gradient of $d$ w.r.t. $\vec{\xi}^j$ can be expressed as
\begin{equation}
\label{eq:signed_dist_grad_1}
\vec{J}_{\vec{\xi}^j}
\triangleq
\frac{\partial d}{\partial \vec{\xi}^j}
=
\operatorname{sign}(d)\,
\frac{{\Delta \vec{p}^*}^\top}{\|\Delta \vec{p}^*\|}\,
\frac{\partial \Delta \vec{p}^*}{\partial \vec{\xi}^j},
\end{equation}
where $\operatorname{sign}(d)$ indicates either separation or penetration.

The formulation above appears to require second-order information of the support functions. 
One possible way to approximate these Hessian-related terms is to use sampling-based gradient-estimation techniques, such as randomized smoothing~\cite{Montaut2023RandomizedSmoothing}. 
However, in the context of SDF, this approximation is not necessary. 
By exploiting the positive homogeneity of the support function, the Hessian-related terms are shown to have no contribution after projection along the separation direction $\Delta \vec p^*$. 
This yields a closed-form expression without Hessian approximation.

\paragraph{Simplified SDF gradient}
We use the fact that the support function $\sigma_{\mathcal Q}(\cdot)$ is positively homogeneous of degree one. 
By Euler's homogeneous function theorem, this implies
$\nabla \sigma_{\mathcal Q}(\vec{z})^\top \vec{z}
= \sigma_{\mathcal Q}(\vec{z})$.
Taking the derivative with respect to $\vec{z}$ leads to
$\nabla^2 \sigma_{\mathcal Q}(\vec{z})\, \vec{z}
= \vec{0}$.
Using this result for $\mathcal Q^k$ with $\vec{z} = - \vec{R}_k^\top \Delta \vec{p}^*$ for $k \in \{i,j\}$ yields
\begin{equation}
\vec{H}_i \, \vec{R}_i^\top \Delta \vec{p}^* = \vec{0},
\quad
\vec{H}_j \, \vec{R}_j^\top \Delta \vec{p}^* = \vec{0}.
\label{eq:simp_1}
\end{equation}
Right-multiplying~\eqref{eq:df_ddp} by $\Delta \vec p^*$ and using
\eqref{eq:simp_1} gives
$\partial \vec f_s / \partial \Delta \vec p^* \, \Delta \vec p^*
=\Delta \vec p^*$.
Since the Hessians of convex SQs are positive semidefinite, 
$\partial \vec{f}_s / \partial \Delta \vec{p}^* = \vec{I} + (\cdot)$ is positive definite and invertible. 
Inverting and transposing gives
\begin{equation}
\label{eq:dpT_Minv}
{\Delta \vec{p}^*}^\top
\Big(\frac{\partial \vec{f}_s}{\partial \Delta \vec{p}^*}\Big)^{-1}
=
{\Delta \vec{p}^*}^\top .
\end{equation}
Invoking~\eqref{eq:implicit_cond} and~\eqref{eq:dpT_Minv} simplifies the SDF gradient~\eqref{eq:signed_dist_grad_1} to
$\vec{J}_{\vec{\xi}^j}
=
-\operatorname{sign}(d)\,
{\Delta \vec{p}^*}^\top/\|\Delta \vec{p}^*\|\,
\left.\partial \vec{f}_s/\partial \vec{\xi}^j\right|_{\Delta \vec{p}^*}$.
Incorporating~\eqref{eq:df_dx} and exploiting~\eqref{eq:simp_1} to further simplify the rotational part yields the final expression of the SDF gradient:
\begin{equation}
\vec{J}_{\vec{\xi}^j}
=
-\operatorname{sign}(d)\,
\frac{{\Delta \vec{p}^*}^\top}{\|\Delta \vec{p}^*\|}\,
\big[\,
\vec{I} \quad 
- \vec{R}_j \, [\vec{p}_j^*]_\times \, \vec{J}_r(\vec{\theta}_j)
\,\big],
\label{eq:signed_dist_grad_final}
\end{equation}
thereby eliminating all computationally intensive Hessian terms while preserving exactness.

\revthree{
Note that, as established in~\autoref{subsec:SdfCbf}, strictly convex SQs ensure that $\vec{J}_{\vec{\xi}^j}$ is continuous on the safe set.
The derivation above further establishes~\eqref{eq:signed_dist_grad_final} on the non-degenerate configurations, which is a dense subset of the safe set.
Both sides of~\eqref{eq:signed_dist_grad_final} are continuous and coincide on this dense subset; hence, this identity extends to the whole safe set for $e_{1,2}\in(0,2)$~\cite{dieudonne1960foundations}.
The $\mathcal{C}^2$ continuity condition on $\sigma_{\mathcal{Q}}$, therefore, should be interpreted as an artifact of the derivation rather than a prerequisite for~\eqref{eq:signed_dist_grad_final} to be applicable.
}

}

\subsection{SDF-CBF Condition}

With the simplified SDF gradient% introduced in~\autoref{subsec:SdfGradEst}
, the CBF time derivative can be efficiently computed in closed form.
Using the chain rule, the SDF's time derivative is
\begin{equation}
\dot{d}^{i,j}(\vec{q}, \vec{\xi}^j_{\mathrm{obs}})
= 
\vec{J}^{i,j}_{\vec{\xi}^i}
\dot{\vec{\xi}}^i_{\mathrm{rob}}(\vec{q}) 
+ 
\vec{J}^{i,j}_{\vec{\xi}^j}
\dot{\vec{\xi}}^j_{\mathrm{obs}},
\label{eq:cbf_derivative}
\end{equation}

\noindent where SDF gradients $\vec{J}^{i,j}_{\vec{\xi}^i} = \partial d^{i,j} / \partial \vec{\xi}^i_{\mathrm{rob}}$ and $\vec{J}^{i,j}_{\vec{\xi}^j} = \partial d^{i,j} / \partial \vec{\xi}^j_{\mathrm{obs}} $ 
can be computed with~\eqref{eq:signed_dist_grad_final}. 
While the time derivative of the obstacle state $\dot{\vec{\xi}}^j_{\mathrm{obs}}$ 
can be directly obtained from the perception pipeline, the state rates of robot SQ
$\dot{\vec{\xi}}^i_{\mathrm{rob}}$ 
must be derived from $\dot{\vec{q}}$ using robot kinematics model. 

Specifically, for the $i$-th robot SQ attached to link $\ell(i)$, 
we map the joint velocities to the SQ's analytical state derivative using a two-stage kinematic transformation. 
First, we map the joint velocities to the spatial twist: 
\begin{equation*}
\begin{bmatrix}
    \vec{v}_{\ell(i)} \\
    \boldsymbol{\omega}_{\ell(i)}
\end{bmatrix} = \vec{J}_{\ell(i)}(\vec{q}) 
\dot{\vec{q}} =
    \begin{bmatrix}
\vec{J}^{v}_{\ell(i)}(\vec{q}) \\[4pt]
\vec{J}^{\omega}_{\ell(i)}(\vec{q})
\end{bmatrix}
\dot{\vec{q}} \,,
\end{equation*}
where $\vec{J}_{\ell(i)}(\vec{q})$ denotes the space Jacobian 
up to link $\ell(i)$~\cite{Lynch2017ModernRobotics}. 
Second, we transform the spatial twist into the analytical state derivative of the attached SQ:
\begin{equation*}
\dot{\vec{\xi}}^i_{\mathrm{rob}}
=
\vec{X}_{i}(\vec{q})
 \begin{bmatrix}
    \vec{v}_{\ell(i)} \\
    \boldsymbol{\omega}_{\ell(i)}
\end{bmatrix} =
\begin{bmatrix}
\vec{I}_3 & -[\vec{t}_{i}]_\times \\[4pt]
\vec{0}_3 & \vec{J}_l^{-1}(\vec{\theta}_{i})
\end{bmatrix}
 \begin{bmatrix}
    \vec{v}_{\ell(i)} \\
    \boldsymbol{\omega}_{\ell(i)}
\end{bmatrix} \,,
\end{equation*}
where %$\left[\cdot\right]_\times$ is the skew-symmetric operator, 
$\vec{X}_{i}(\vec{q})$ maps the spatial twist to the pose rate of geometry $i$, and $\vec{J}_l(\cdot)$ is the left Jacobian of $\mathrm{SO}(3)$~\cite{Joan2018MircoLie}.

By choosing $\vec{u} = \dot{\vec{q}}$, the SDF-CBF condition for collision avoidance between a robot SQ and an obstacle SQ can be compactly written as
$\vec{J}^{i, j}_{\delta|i}
\vec{u} + 
\dot{d}^{i,j}_{\mathrm{obs}}
\ge - \alpha_{\delta}(h^{i,j}_{\delta}
),  \quad 
(i,j) \in \mathcal{P}_{\text{env}}$,
where
$\vec{J}^{i, j}_{\delta|i}
= 
\vec{J}^{i,j}_{\vec{\xi}^i}
\vec{X}_{i}
\vec{J}_{\ell(i)}$, 
$\dot{d}^{i,j}_{\mathrm{obs}}
= 
\vec{J}^{i,j}_{\vec{\xi}^j}
\dot{\vec{\xi}}^j_{\mathrm{obs}}$, and the index set is defined as
$\mathcal{P}_{\text{env}} = \mathcal{I}^{\mathrm{rob}} \times \mathcal{I}^{\mathrm{obs}}$.
The function $\alpha_\delta(\cdot)$ denotes an extended class-$\mathcal{K}$ function.
\revone{
For self-collision avoidance, the formulation can be readily extended to
$\big(\vec{J}^{i, j}_{\delta|i} + \vec{J}^{i, j}_{\delta|j}\big)
\vec{u}
\ge - \alpha_{\delta}(h^{i,j}_{\delta}
),  \quad 
(i,j) \in \mathcal{P}_{\text{self}}$,
where $\mathcal{P}_{\text{self}} \subset \mathcal{I}^{\mathrm{rob}} \times \mathcal{I}^{\mathrm{rob}}$. Note that collision checks are excluded for pairs with $i = j$ and for adjacent links.
}

\subsection{Safety Filter Formulation}

To improve task-space command tracking~\cite{Daniel2025OSCBF}, we augment the objective~\eqref{eq:cbf_qp} with a task-consistency term that penalizes deviations in task-space, i.e., $\| \vec{J}(\vec{q}) ( \vec{u} - \vec{u}_{\mathrm{cmd}} ) \|_2^2$.
The proposed safety filter can be formulated as
\begin{align}
  \vec{u}^* = 
   \; \arg&\min_{\vec{u} \in \mathcal{U}}  
  \big\|\vec{u} - \vec{u}_{\mathrm{cmd}}\big\|^2_2 +
  \big\|\vec{J}(\vec{q}) (\vec{u} - \vec{u}_{\mathrm{cmd}})\big\|^2_2 \label{eq:final_formulation}\\
  \text{s.t.} 
  & \; \vec{J}^{i,j}_{\delta|i} \vec{u}  + 
    \dot{d}^{i,j}_{\mathrm{obs}} 
    \ge - \alpha_{\delta}(h^{i,j}_{\delta}), \quad
    \forall (i, j) \in \mathcal{P}_{\text{env}} \,,
    \nonumber\\
  & \; \big(\vec{J}^{i,j}_{\delta|i} + \vec{J}^{i,j}_{\delta|j} \big)\vec{u}
    \ge - \alpha_{\delta}(h^{i,j}_{\delta}), \quad
    \forall (i, j) \in \mathcal{P}_{\text{self}} \,,
    \nonumber
\end{align}
where we have dropped function arguments for brevity. 
\revone{Since all constraints are affine in $\vec{u}$,
problem~\eqref{eq:final_formulation} can be solved efficiently online using a standard QP solver.}

\revone{
\section{Implementation and Robustness Analysis}
\label{sec:robustness}

Our formulation assumes access to the exact SDF between SQs. 
However, this requires solving a constrained optimization problem over implicitly defined surfaces, which is computationally expensive. 
We therefore approximate each SQ by a convex polytope $\mathcal{P}$ obtained via uniform surface sampling, where $n_u$ and $n_v$ denote the sampling resolutions along the longitudinal and latitudinal coordinates $u$ and $v$. 
This enables fast SDF evaluation via GJK/EPA~\cite{GJK1988,VanDenBergen2004EPA}.

While this discretization significantly improves computational efficiency, it introduces geometric approximation errors that affect both the SDF and its time derivative used in the CBF condition. In the following, we quantify these errors and derive a principled robustness formulation.
The geometric error of the $k$-th SQ's surface with area $A_k$ can be decomposed into tangential and normal components $\delta_{T,k}$ and $\delta_{N,k}$, respectively.
Assuming quasi-uniform equal-area sampling using local hexagonal patches with area $a_k = A_k / (n_{u}  n_{v})$,
the tangential error is 
$\delta_{T,k}=\sqrt{c_{a,k}} \, r_{\text{hex},k}$
where $r_{\text{hex},k}$ denotes the hexagon's circumradius and $c_{a,k}\ge 1$ accounts for sampling non-uniformity. 
Using a second-order approximation of the exact circular sagitta with a curvature radius lower bounded
by $\rho_k$, we get
$\delta_{N,k} =
\delta_{T,k}^2 / (2\rho_k)$.

Since $\mathcal{P}_k \subset \mathcal{Q}_k$ for $k \in \{ i, j\}$, the support-function gap is $0\le
\sigma_{\mathcal Q_k}(\vec n)
-
\sigma_{\mathcal P_k}(\vec n)
\le
\delta_{N,k}, \forall \vec n\in\mathbb S^2$. 
Let $\varphi(\vec n)
:=
-\sigma_{\mathcal Q_i}(\vec n)
-\sigma_{\mathcal Q_j}(-\vec n)$ and $\tilde{\varphi}(\vec n)$ be defined analogously over $\mathcal P_i$, $\mathcal P_j$. 
For separated convex bodies, the Euclidean set distance admits the support-function representation
$d^{i,j}=\max_{\vec n \in \mathbb{S}^2}\varphi(\vec n),$ and $
\tilde d^{i,j}=\max_{\vec n \in \mathbb{S}^2}\tilde \varphi(\vec n)$. %~\cite{Jame2022Lemma}. 
Moreover, the inner approximation gives 
$\tilde d^{i,j} \ge d^{i,j}$. 
Applying the support-function gap to body $i$ along $\vec n$ and to body $j$ along
$-\vec n$ gives the pointwise relation 
$\varphi(\vec n)
\le
\tilde\varphi(\vec n)
\le
\varphi(\vec n)+\delta_d^{i,j},
$ with $
\delta_d^{i,j}:=\delta_{N,i}+\delta_{N,j}$.
Taking the maximum over $\vec n$ yields the SDF error bound $0 \le \tilde d^{i,j}-d^{i,j} \le \delta_d^{i,j}$.

Let $\Delta \tilde{\vec p}^*$
denote the separation between $\mathcal{P}_i$ and $\mathcal{P}_j$ and let $\vec n = \Delta \vec p^* / \|\Delta \vec p^*\|$, and $\tilde{\vec n} = \Delta \tilde{\vec p}^*/ \|\Delta \tilde{\vec p}^*\|$
denote the exact and discretized separation directions, where $\|\Delta \vec p^*\| = d^{i,j}$ and $\|\Delta \tilde{\vec p}^*\| = \tilde d^{i,j}$. The supporting hyperplane separating $\mathcal Q_i$ from $\mathcal Q_j$
along $\vec n$ also separates the inscribed polytopes, so the
discretized separation projects onto $\vec n$ with $\vec n^\top \Delta \tilde{\vec p}^* \ge d^{i,j}$. Therefore $\vec n^\top \tilde{\vec n}
= \vec n^\top \Delta \tilde{\vec p}^* / \tilde d^{i,j} \ge d^{i,j} / (d^{i,j} + \delta_d^{i,j})$.
Using the unit-vector identity
$\|\tilde{\vec n} - \vec n\|^2 = 2(1 - \vec n^\top \tilde{\vec n})$ and
$d^{i,j} \ge \underline\epsilon_\delta$, the minimum exact clearance, we obtain $\|\tilde{\vec n} - \vec n\| \le \beta^{i,j}$ where $\beta^{i,j} = \sqrt{2\,\delta_d^{i,j} / \underline\epsilon_\delta}$.
This direction error propagates through the SDF Jacobian in \eqref{eq:cbf_derivative}:
the translational Jacobian error
is bounded by $\beta^{i,j}$ and the rotational one by
$\bar r_k\beta^{i,j} + \delta_k$, with $\bar r_k$ an upper bound on
$\|\vec p_k^*\|$ in $\mathcal F^k$, based on the encapsulating cuboid's space diagonal. Combined with the velocity bounds
$\|\dot{\vec t}^{\,k}\| \le \bar v_k$ and
$\|\dot{\vec\theta}^{\,k}\| \le \bar\omega_k$, the SDF derivative
error satisfies $|\tilde{\dot d}^{i,j} - \dot d^{i,j}| \le \gamma^{i,j}$
with $\gamma^{i,j}
= \beta^{i,j}(\bar v_i + \bar v_j)
+ (\bar r_i\beta^{i,j} + \delta_i)\bar\omega_i
+ (\bar r_j\beta^{i,j} + \delta_j)\bar\omega_j$.

\revthree{
Note that the theoretical error bounds derived above from the SQ geometry and sampling resolution can be conservative.
Alternatively, tighter empirical estimates of the SDF and gradient errors can be obtained from the SQ configurations recorded along the experimental trajectories.
For each configuration, we compute a reference distance by solving the constrained distance minimization problem over the continuous SQ surfaces, as in~\autoref{fig:inside-outside}, and obtain reference gradients using central finite differences.
The maximum difference between the reference and discretized
distances gives the empirical SDF error estimate $\delta_d^{i,j}$, and the maximum gradient errors are combined with the velocity bounds
to estimate the SDF time-derivative error $\gamma^{i,j}$.
}

The discretization errors can now be compensated by tightening the CBF margin. 
This parallels input-to-state safe CBFs~\cite{Kolathaya2019ISSfCBF}, except that the bounded term is the polytopic approximation error rather than an external disturbance.
By monotonicity of $\alpha_\delta$, the exact CBF condition
$\dot d^{i,j} \ge -\alpha_\delta(d^{i,j} - \underline\epsilon_\delta)$
is implied by $\tilde{\dot d}^{i,j}
\ge
-\alpha_\delta\!(
\tilde d^{i,j} - \underline\epsilon_\delta - \delta_d^{i,j}
)
+
\gamma^{i,j}$.
For the linear choice $\alpha_\delta(\rho) = \kappa_\delta\rho$ with $\kappa_\delta > 0$,
this robust condition rearranges into the standard CBF form $\tilde{\dot d}^{i,j}
\ge
-\kappa_\delta\!\left(
\tilde d^{i,j} - \epsilon_\delta^{i,j}
\right)$, with the tightened safety margin $\epsilon_\delta^{i,j}
=
\underline\epsilon_\delta
+
\Delta\epsilon_\delta^{i,j}$, where $\Delta\epsilon_\delta^{i,j}
=
\delta_d^{i,j}
+ \gamma^{i,j}/ \kappa_\delta$. Therefore, enforcing the CBF constraint on the discretized polytopic with margin $\epsilon_\delta^{i,j}$ guarantees the exact CBF
condition for the original SQs.
\revthree{
In practice, this error is often relatively small; for instance, for our evaluated experimental configurations, the empirically estimated upper bound is $\le\SI{1}{\milli\meter}$.
}
}

\begin{table*}[t]
\centering
\caption{Comparison of geometric collision models. Bold entries are from our proposed method, while notable limitations are highlighted in red. 
}
\label{tab:case_comparison}

\renewcommand{\arraystretch}{1.0}        % increase row height
\setlength{\tabcolsep}{10pt}               % column padding

\begin{tabular}{c|c c c|c}

\toprule[1.2pt]

Representation
& Sphere~\cite{Daniel2025OSCBF}
& Cylinder / Box~\cite{Vinicius2024SmoothDistances}
& \textbf{Superquadric (ours)} 
& Official FR3 Collision Mesh \\

\midrule

Illustration
&
\begin{minipage}[b]{0.25\columnwidth}
\centering
\vspace*{1mm}
\raisebox{-.5\height}{
\includegraphics[
            width=\linewidth,
            trim=20 85 20 70,   % left bottom right top
            clip
        ]{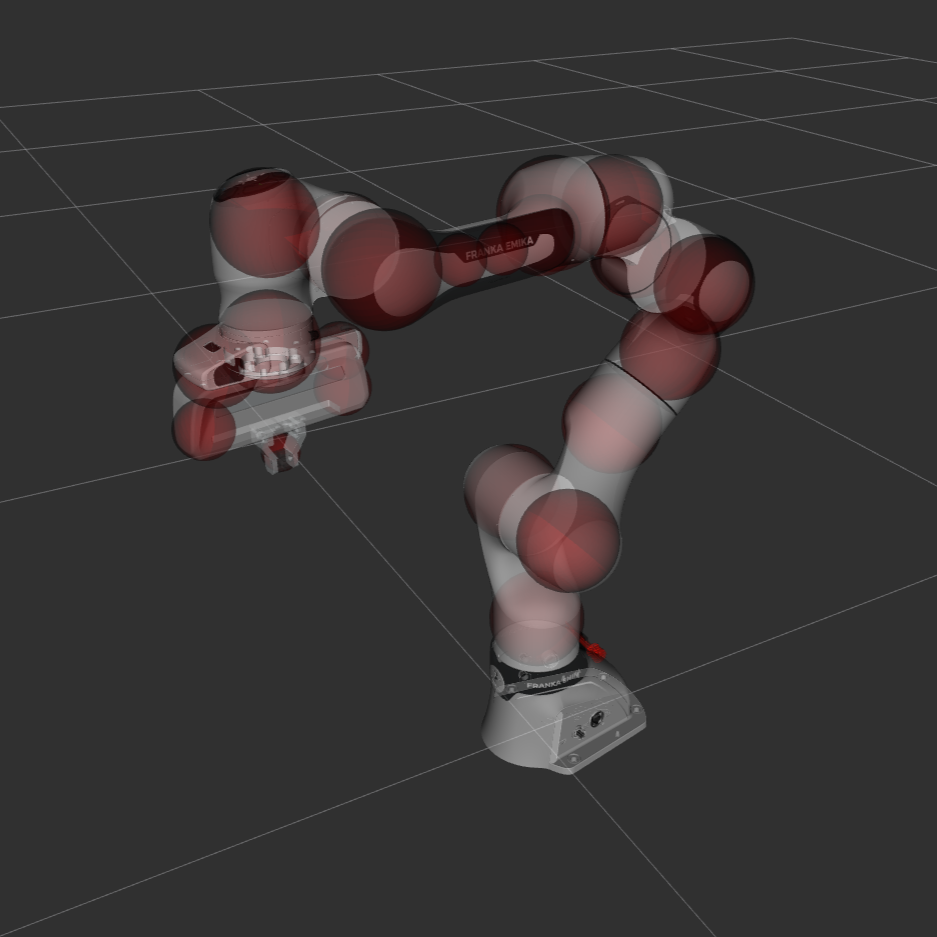}
}
\end{minipage}
& 
\begin{minipage}[b]{0.25\columnwidth}
\centering
\vspace*{1mm}
\raisebox{-.5\height}{
\includegraphics[
            width=\linewidth,
            trim=20 85 20 70,   % left bottom right top
            clip
        ]{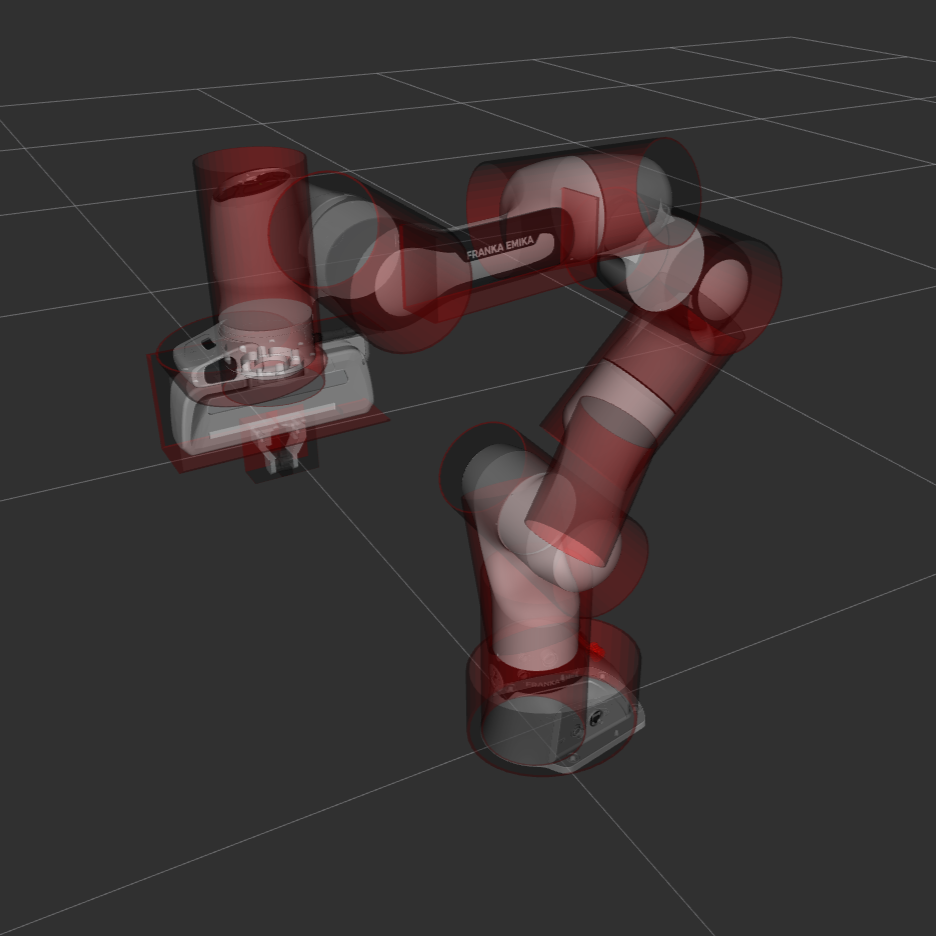}
}
\end{minipage}
&
\begin{minipage}[b]{0.25\columnwidth}
\centering
\vspace*{1mm}
\raisebox{-.5\height}{
\includegraphics[
            width=\linewidth,
            trim=20 85 20 70,   % left bottom right top
            clip
        ]{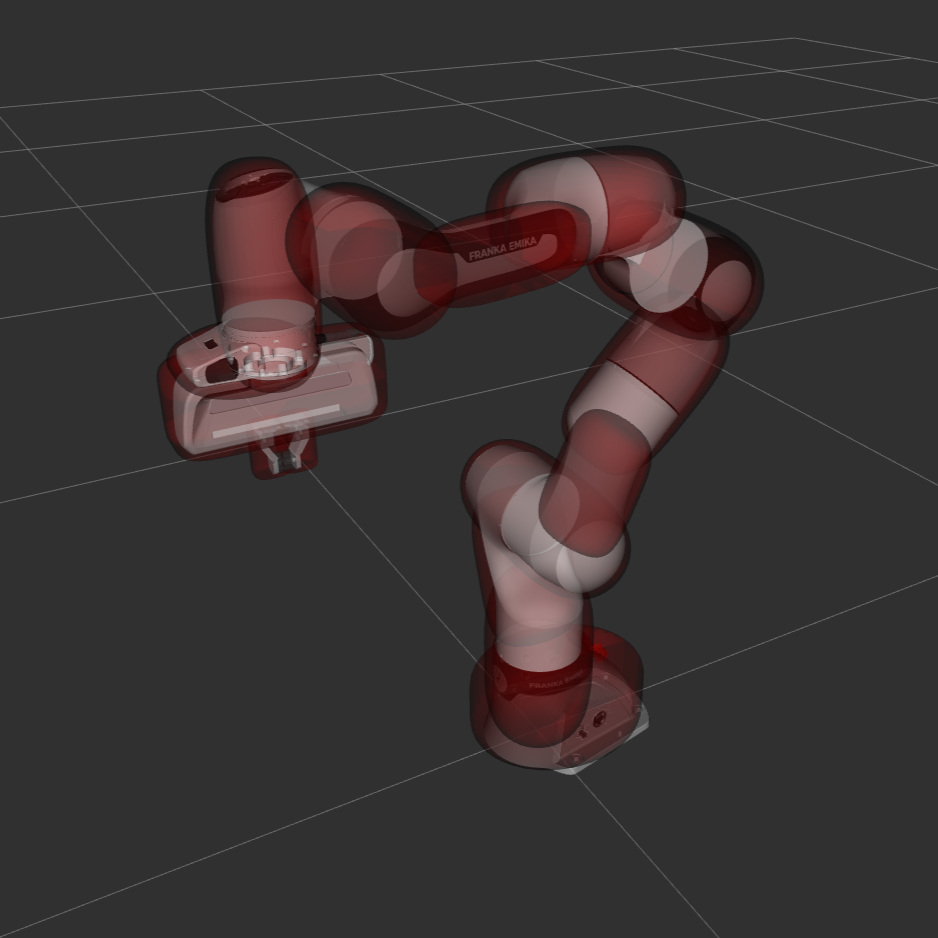}
}
\end{minipage}
& 
\begin{minipage}[b]{0.25\columnwidth}
\centering
\vspace*{1mm}
\raisebox{-.5\height}{
\includegraphics[
            width=\linewidth,
            trim=20 85 20 70,   % left bottom right top
            clip
        ]{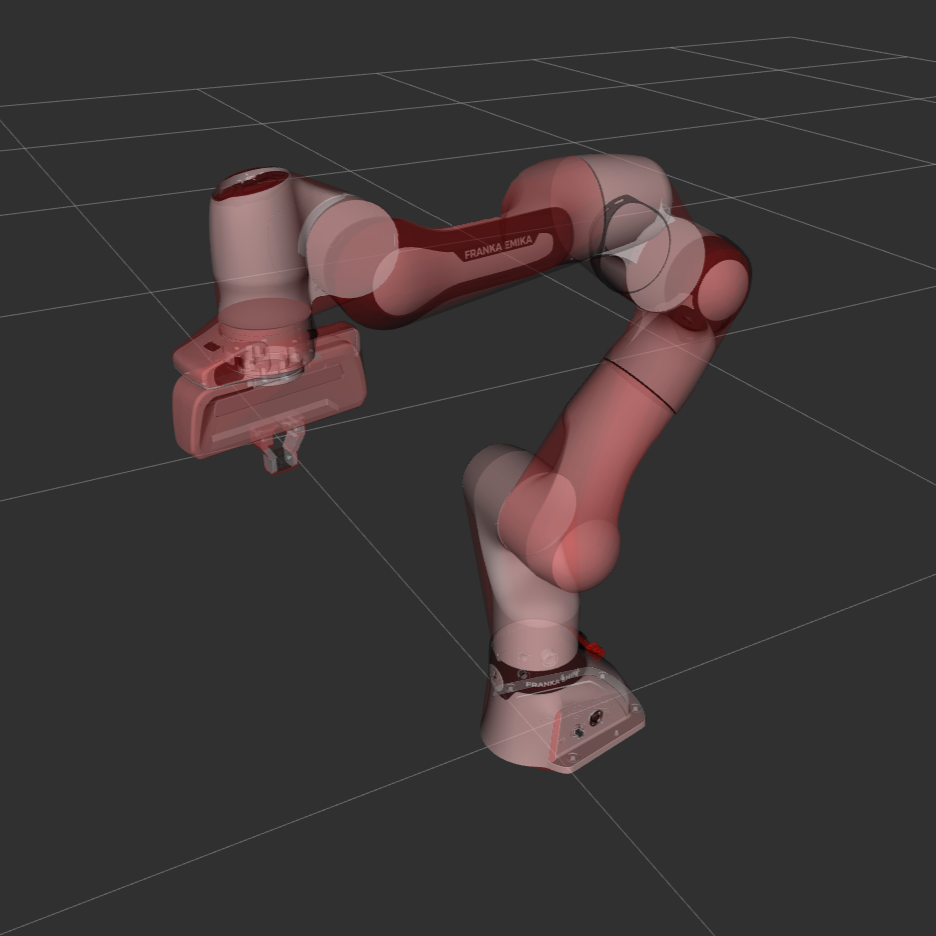}
}
\end{minipage} \\

\rule{0pt}{3ex}Geometry Count
& 21 Spheres
& 11 Cylinders + 4 Boxes
& \textbf{15 Superquadrics}
& 9 Tri. Meshes + 8 Boxes\\ 

\rule{0pt}{3ex}Coverage ($\uparrow$)
& \cellcolor{red!20} 59.08\%
& 99.00\%
& \textbf{97.14\%}
& 97.28\% \\ 

\rule{0pt}{3ex}OverApprox. ($\downarrow$)
& 2.57\%
& \cellcolor{red!20} 45.23\%
& \textbf{28.69\%}
& 1.92\% \\

\rule{0pt}{3ex}Convexity
& strictly convex
& \cellcolor{red!20} not strictly convex$^{\dagger}$
& \textbf{strictly convex$^{\ddagger}$}
& \cellcolor{red!20} non-convex \\

\bottomrule[1.2pt]

\end{tabular}

\vspace{1mm}
\footnotesize
\noindent
$^{\dagger}$~A cylinder or a box is not strictly convex, since it contains nontrivial line segments.
\quad
$^{\ddagger}$~A superquadric is strictly convex when $e_{1,2} \le 2$.
\vspace*{-2em}
\end{table*}

\section{Experiments}
\label{sec:expr-section}

In this section, we present the experimental evaluation of the proposed safety filter. 
We begin by addressing two key questions: 
\textit{(i)} how expressive and compact the proposed SQ collision model is, and 
\textit{(ii)} whether SDF gradients can be efficiently obtained online for reliable safety filtering. 
We then present simulation and real-world results demonstrating the safety filter's behavior in manipulation tasks.

\subsection{Implementation Details}
\label{subsec:implementation_details}

All evaluations and experiments are conducted on a workstation equipped with an Intel Core Ultra 9 285K CPU, \revone{using a Franka Emika FR3 robot with joint velocity control $\vec{u} \in \mathbb{R}^7$.}
The pairwise distances are returned by the GJK/EPA algorithm implemented in the Coal library~\cite{coal}.
Each SQ's surface is uniformly sampled with $n_u = n_v = 200$.
For the SDF-CBF, we choose $\kappa_\delta = 2$ and set the 
minimal clearance $\underline\epsilon_\delta = \SI{10}{\milli\meter}$. 
\revthree{
To account for surface discretization error, we evaluate the robustified safety margin using the two largest SQs with $\vec{s}_1=[0.11,0.05,0.03,0.3,0.3]^\top$ and $\vec{s}_2=[0.145,0.065,0.105,0.3,0.3]^\top$, and use maximum translational and angular velocities of $\bar v_k=\SI{0.10}{\meter\per\second}$ and $\bar\omega_k=\SI{0.33}{\radian\per\second}$, respectively.
The empirically estimated maximum errors in the SDF and its
time derivative are
$\delta_d^{i,j}\approx\SI{1.10e-5}{\meter}$ and
$\gamma^{i,j}\approx\SI{1.7e-3}{\meter\per\second}$, yielding an additional safety margin of
$\Delta\epsilon_\delta^{i,j}\approx\SI{0.84}{\milli\meter}$.
}
In the real-world experiments, we additionally penalize input discontinuity using $w \|\vec{u}_k - \vec{u}_{k-1}\|^2_2$ with $w = 0.1$.
The CBF-QP~\eqref{eq:final_formulation} is solved in real time at $\SI{100}{\hertz}$.

\subsection{Collision Model Representation}

We compare the proposed SQ-based model with: \textit{(i)} sphere-based model~\cite{Daniel2025OSCBF}
and \textit{(ii)} cylinder-box model~\cite{Vinicius2024SmoothDistances}, while the official FR3 collision model is also included for reference.
The comparison considers: \textit{(i)} number of primitives, \textit{(ii)} coverage and over-approximation, and \textit{(iii)} convexity. 

We denote the true robot geometry by $\mathcal{R} \subset \mathbb{R}^3$ and the fitted collision model by $\mathcal{C} \subset \mathbb{R}^3$. 
The coverage and over-approximation ratios are computed using a voxel discretization.
Let $\Omega \subset \R^3$ be a region that contains $\mathcal{R} \cup \mathcal{C}$. A uniform discretization of $\Omega$ with resolution $\Delta = \SI{0.005}{\meter}$ yields the voxel set $\Omega_\Delta$.  
Then the voxelized robot and collision model geometries are $\mathcal{V}_{\mathcal{R}} = \{v_i \in \Omega_\Delta \mid v_i \cap \mathcal{R} \neq \emptyset\}$ and
$\mathcal{V}_{\mathcal{C}} = \{v_i \in \Omega_\Delta \mid v_i \cap \mathcal{C} \neq \emptyset\}$, respectively.
The coverage and over-approximation ratios are defined as
$\mathrm{Coverage} =
|\mathcal{V}_{\mathcal{R}} \cap \mathcal{V}_{\mathcal{C}}|/|\mathcal{V}_{\mathcal{R}}|$ and
$\mathrm{OverApprox} = |\mathcal{V}_{\mathcal{C}} \setminus \mathcal{V}_{\mathcal{R}}|/|\mathcal{V}_{\mathcal{R}}|$.

The comparison of collision models is summarized in~\autoref{tab:case_comparison}.
The sphere-based model is least conservative but provides less than $\!60\%$ coverage, as spheres poorly capture thin structures and sharp features, leading to missed collisions (see \autoref{subsec:sim}).
The cylinder-box decomposition achieves near-complete coverage but incurs almost $\!50\%$ over-approximation and requires additional smoothing for gradient-based optimization~\cite{Vinicius2024SmoothDistances}.
In contrast, the proposed SQ-based model balances coverage and conservativeness: 
it achieves coverage comparable to the official FR3 collision model, reduces over-approximation relative to the cylinder-box approach, and uses fewer geometries.

\begin{figure}[t]
    \centering
        \includegraphics[width=\linewidth]{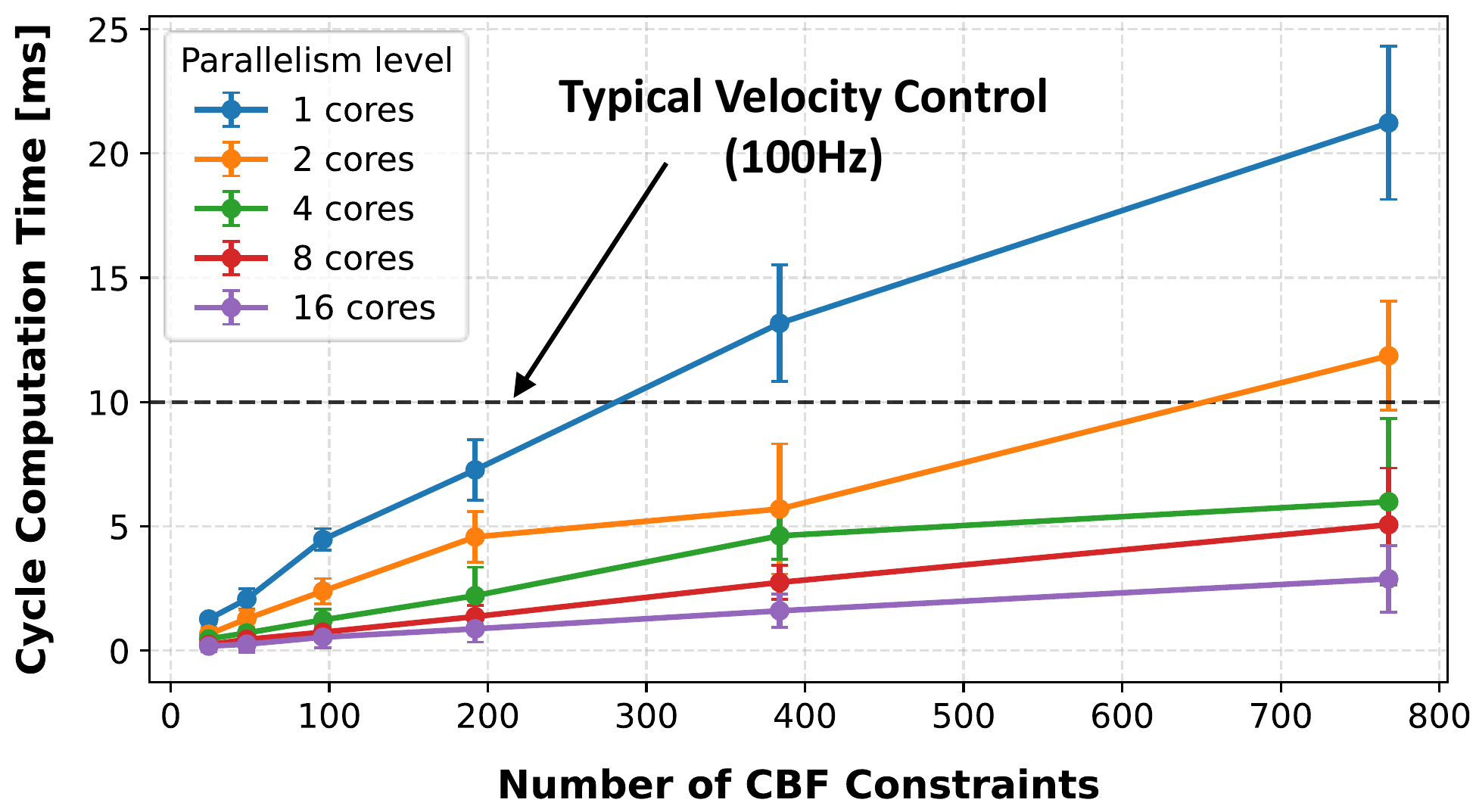}
        \caption{\revone{Cycle time (mean~$\pm$~std.) of the proposed SDF and gradient evaluation pipeline as a function of constraint number under different CPU parallelization levels. Multi-core execution significantly improves scalability.}}
    \label{fig:cycle_time}
\end{figure}

\subsection{Computational Efficiency and Approximation Accuracy}

\begin{figure}[t]
    \centering
        \includegraphics[width=1.0\linewidth]{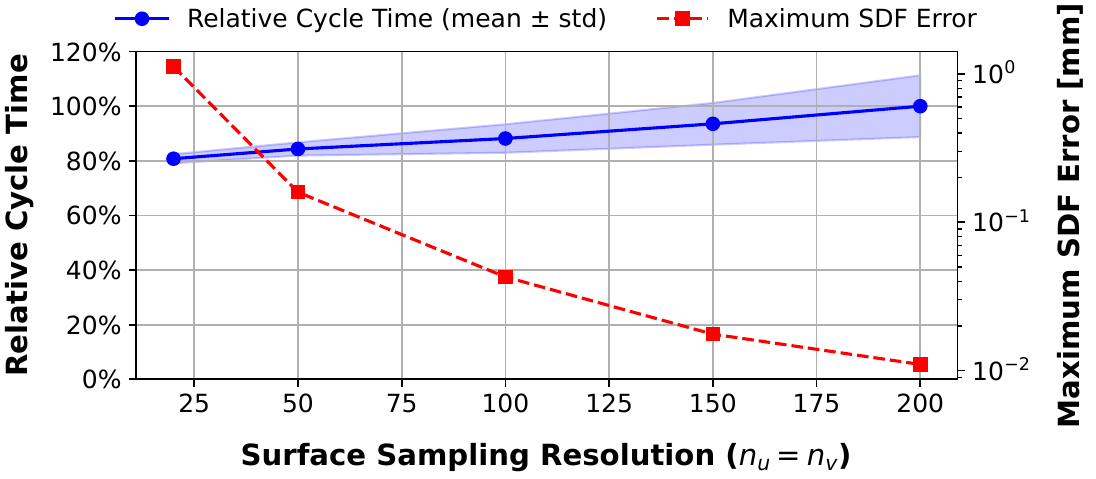}
        \caption{\revone{Trade-off between efficiency and accuracy with increasing resolution $n_u = n_v$. 
        The blue curve shows the relative cycle time normalized w.r.t. the $200 \times 200$ resolution, while the red dashed curve (log-scale) shows the empirical SDF error. Increasing the resolution significantly reduces discretization error by approximately two orders of magnitude, at the cost of slightly increased computation time.}}
    \label{fig:runtime_error_tradeoff}
\end{figure}

\autoref{fig:cycle_time} reports the cycle time versus the number of CBF constraints under different parallelization levels.
On a single CPU core, the cycle time grows approximately linearly with the number of CBF constraints, and still \revone{supports real-time velocity control for more than $200$ collision pairs}. 
For larger numbers of collision pairs, maintaining low cycle times requires exploiting parallel computation. 
With $16$ CPU cores, our \revone{method evaluates about $800$ collision pairs in $2.7\,\mathrm{ms}$}, indicating that the proposed method scales well beyond the scene typically encountered in real-world.
\revone{\autoref{fig:runtime_error_tradeoff} further illustrates the trade-off between efficiency and accuracy: as the sampling resolution increases from $20$ to $200$, the relative cycle time increases only from $80\%$ to $100\%$, while the maximum SDF error decreases by two orders of magnitude. 
This indicates that finer meshes substantially improve approximation accuracy at a modest computational cost.
}

\subsection{Simulation Experiments}
\label{subsec:sim}

\begin{figure}[t]
    \centering
        \includegraphics[width=0.98\linewidth]{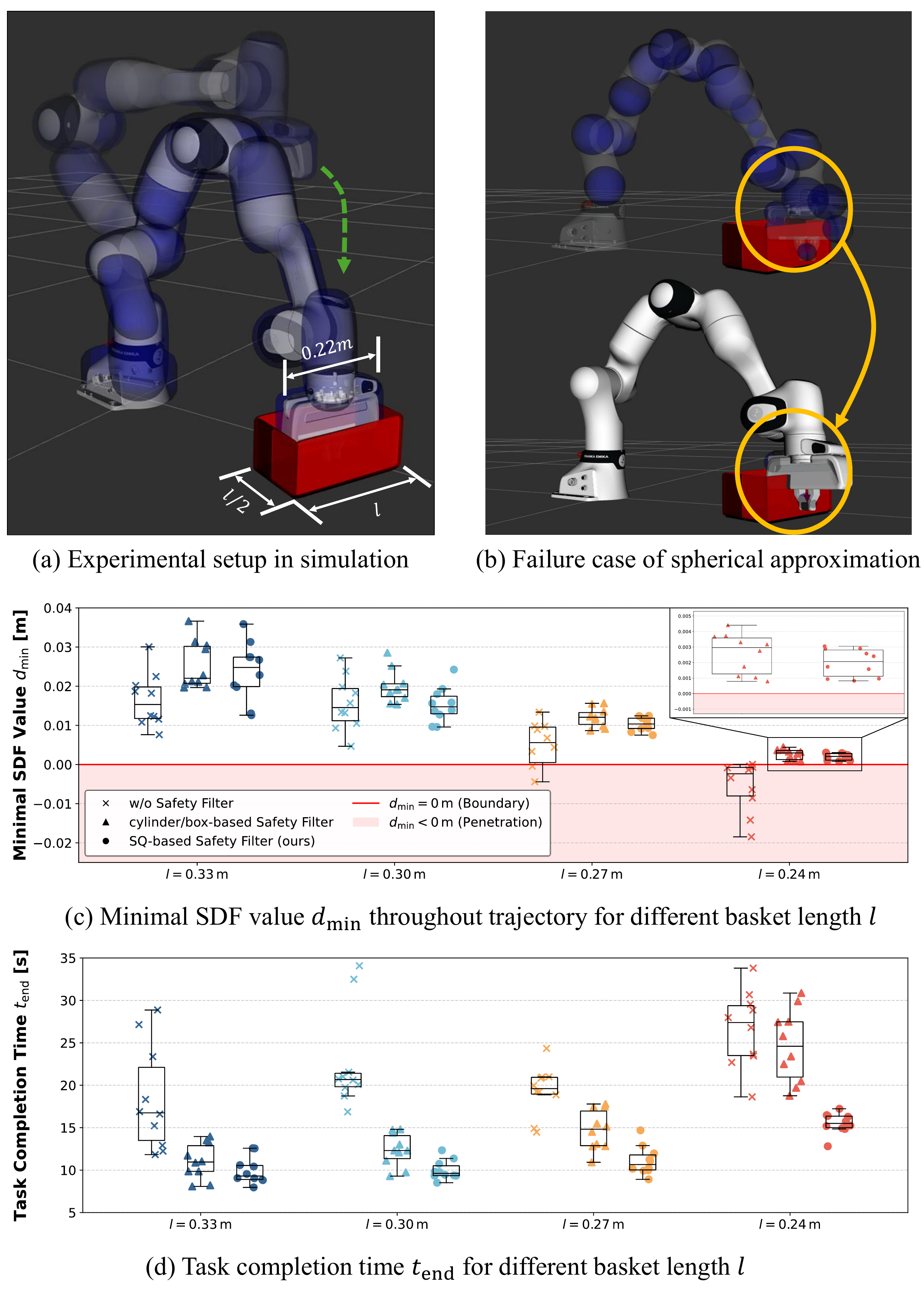}
        \caption{\revtwo{Simulation experiment of a teleoperated insertion task under progressively tighter geometric constraints.  
        The results show that the proposed superquadric-based safety filter consistently prevents collisions and reduces the task completion time by $47.5\%$ on average compared with the baseline without a safety filter, and by $26.8\%$ on average compared with the cylinder/box-based safety filter. In contrast, the coarse spherical model from~\cite{Daniel2025OSCBF} may be insufficient to guarantee safety.}}
    \label{fig:sim_exp}
\end{figure}

We evaluate the proposed safety filter in simulation to assess its safety and efficiency in challenging environments, particularly when the robot must move in close proximity to obstacles and overly conservative behaviors may degrade task performance.
As shown in~\hyperref[fig:sim_exp]{\autoref*{fig:sim_exp}a}, the robot performs a teleoperated insertion into a basket-like container with side lengths $l$ and $l/2$. 
Decreasing $l$ reduces clearance and increases task difficulty. 
In the most challenging case ($l=\SI{0.24}{\meter}$), the one-sided end-effector clearance is only $\SI{0.005}{\meter}$. 
To maintain feasibility, we reduce the safety margin to $\underline{\epsilon}_\delta=\SI{0.005}{\meter}$ and $\SI{0.0025}{\meter}$ for the second-to-last and last configurations, respectively. 
Safety is quantified by the minimum SDF $d_{\min}$ along each trial ($d_{\min}<0$ indicates penetration), and efficiency by the task completion time $t_{\text{end}}$.
Each configuration was evaluated over ten teleoperated trials with and without the safety filter, using fixed random seeds to ensure identical obstacle placements for fair comparison.

\hyperref[fig:sim_exp]{\autoref*{fig:sim_exp}b} shows a representative failure case using a spherical approximation~\cite{Daniel2025OSCBF}, where insufficient geometric fidelity leads to collisions. 
In contrast,~\hyperref[fig:sim_exp]{\autoref*{fig:sim_exp}c} shows that the proposed SQ-based safety filter maintains strictly positive SDF values across all tested configurations, whereas executions without the safety filter frequently penetrate obstacles in highly constrained settings. 
\revtwo{Moreover, safety is achieved without sacrificing efficiency: as shown in~\hyperref[fig:sim_exp]{\autoref*{fig:sim_exp}d}, the proposed filter reduces the task completion time by $47.5\%$ on average compared with the baseline~(no safety filter), and by $26.8\%$ on average compared with the cylinder/box-based safety filter.} This improvement arises because users no longer need to repeatedly pause and adjust the end-effector. Instead, collision avoidance is handled online by the safety filter, enabling smoother and more direct task execution.

\subsection{Real-World Experiments}

To validate the proposed safety filter in real-world settings, we design three representative manipulation tasks covering different sources of geometric complexity:
\textit{(i)} object handover with whole-body collision avoidance
in tight spaces,
\textit{(ii)} object transportation with end-effector collision
avoidance in unstructured table-top scenes, and
\textit{(iii)} manipulation with reactive avoidance of dynamic obstacles.

We deploy the proposed safety filter in closed loop to prevent unsafe commands from a non-expert teleoperator, with low-level robot control handled by CRISP~\cite{CRISP2025}. 
For semi-static scenes, a wrist-mounted Femto Bolt RGB-D camera captures scene-level point clouds, which are further segmented at the object level~\cite{Benni2025}. 
These partial observations are registered offline to known complete object models using trimmed ICP~\cite{TrICP2002}. 
For dynamic scenes, obstacle poses are obtained online via motion capture \revone{at $\SI{200}{\hertz}$}, and an extended Kalman filter estimates their velocities for reactive collision avoidance \revone{with a cycle time of $1.72 \pm 0.27$~\si{\milli\second}}.

\begin{figure}[t]
    \centering
        \includegraphics[width=\linewidth]{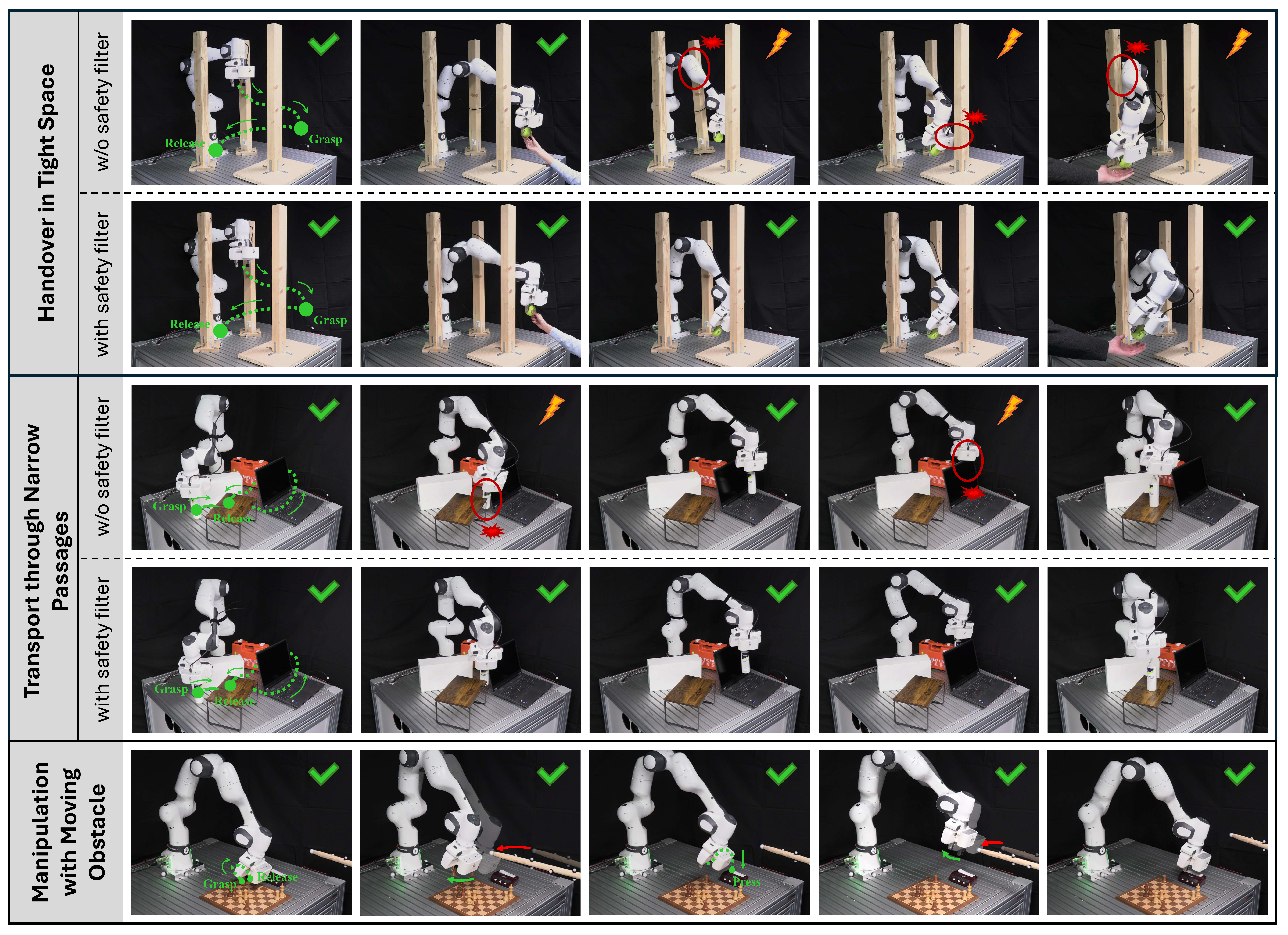}
        \caption{Real-world teleoperated manipulation experiments demonstrating the efficacy of our proposed safety filter on three representative tasks. Across all tasks, the safety filter consistently enables collision-free execution with minimal intervention, despite tight geometric constraints, unstructured and cluttered obstacles, and dynamic disturbances. A video demonstrating the safety filter's performance can be found here: \href{http://tiny.cc/sq-cbf}{http://tiny.cc/sq-cbf}.}
    \label{fig:real_exp}
\end{figure}

\begin{figure}[t]
    \centering
        \includegraphics[width=\linewidth]{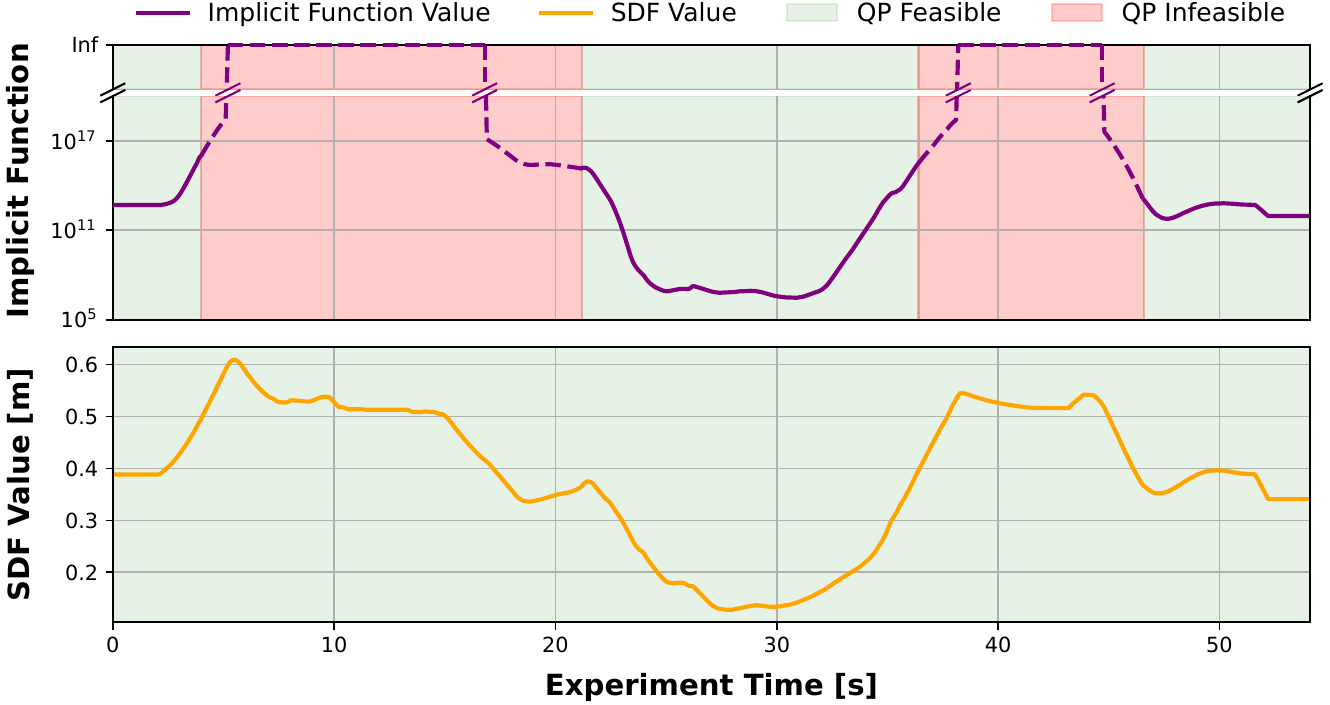}
        \caption{\revone{Comparison of the implicit function-based and the SDF-based formulations along a replayed trajectory from Experiment~2. 
        The implicit function value exhibits extreme variations spanning $10^{5\sim17}$, and occasionally returns numerically saturated values (displayed as $\mathrm{Inf}$) due to the ill-conditioned optimization. These effects lead to infeasible QP instances. In contrast, the SDF remains well-behaved and maintains feasibility throughout.}}
    \label{fig:feasibility_comparison}
\end{figure}

Representative trials from each task are depicted in~\autoref{fig:real_exp} and can be found in~\href{http://tiny.cc/sq-cbf}{http://tiny.cc/sq-cbf}.
The safety filter enables collision-free execution across all tasks despite tight geometric constraints, dynamic disturbances, and sensing noise, demonstrating the robustness and practical applicability of the proposed approach in real-world settings.
\revone{
Since solving the implicit-function-based formulation takes $57.42\,\mathrm{ms}$ per collision pair, 
we evaluate it offline by replaying the successful SDF-based trials. 
As shown by the representative comparison from Experiment~2 in~\autoref{fig:feasibility_comparison}, the implicit formulation produces values spanning approximately $10^5$ to $10^{17}$ and occasionally returns numerically saturated values, which lead to infeasible QP instances.
In contrast, the SDF remains well-conditioned and maintains QP feasibility throughout the trial. 
As summarized in~\autoref{tab:real_world_statistics}, the proposed SDF-based safety filter achieves $100\%$ success over $10$ trials each in real-world Experiments~1 and~2, eliminating the numerical infeasibility observed in the implicit-function baseline~\cite{Lukas2025SemanticallySafety}.
\revtwo{Compared with the DCOL-based baseline~\cite{Dai2023RAL}, our proposed SDF-based filter achieves a higher task success rate in both experiments, while allowing closer yet still positive clearances and substantially shorter completion times.}
\revthree{
The observed performance difference is mainly due to the collision geometry and the SDF-based CBF formulation. The SQ model reduces geometric conservatism while preserving strict convexity; the SDF-based CBF directly specifies the safety margin as Euclidean distance in relevant regions. The geometry-dependent anisotropic clearance induced by DCOL scaling margin, on the other hand, can  complicate safety-margin tuning and  introduce directional conservatism, reducing free space for task execution.
}
These results validate our central claim that the proposed SDF-based SQ-CBF enables efficient, and numerically robust safety filtering.
}

\begin{table}[t]
\centering
\caption{\revtwo{\revone{Evaluation statistics for Experiments~1 and~2 over $10$ trials each. Success rate~(SR) denotes the fraction of successful executions. The minimum SDF is reported as mean~$\pm$~std., with positive values indicating clearance and negative values indicating penetration. Infeasible Frequency denotes the average percentage of CBF-QP infeasible instances during each trial. The best value for each metric is highlighted in blue.}}}
\label{tab:real_world_statistics}
{
\renewcommand{\arraystretch}{1.05}
\setlength{\tabcolsep}{5.7pt}
\resizebox{\linewidth}{!}{
\begin{tabular}{cccccc}
\toprule
\textbf{Task} & 
\textbf{Safety Filter} & 
\textbf{SR~($\uparrow$)} & 
\makecell[c]{\textbf{Minimum}\\ \textbf{SDF ($\geq 0$)}} &
\makecell[c]{\textbf{Infeasible}\\ \textbf{Freq.~($\downarrow$)}} &
\revtwo{\makecell[c]{\textbf{Completion}\\ \textbf{Time$^{\dagger}$~($\downarrow$)}}} \\
\midrule
\multirow{4}{*}{Exp.1} 
       & w/o SF      & $0\%$  & $-82.9 \pm 7.7$~\si{\milli\meter}      & --   & \revtwo{--} \\
       & \revtwo{DCOL-based~\cite{Dai2023RAL}} & \revtwo{$40\%$}  & \revtwo{$21.2 \pm 2.3$~\si{\milli\meter}} & \revtwo{$0\%$} & \revtwo{$29.1 \pm 2.4 s$} \\
       & Impl.-based$^{\ddagger}$~\cite{Lukas2025SemanticallySafety} &  -- & -- & $27.3 \pm 5.3\%$ & \revtwo{--} \\
       & \cellcolor{cyan!20} \textbf{SDF-based (ours)}  & \cellcolor{cyan!20} $\mathbf{100\%}$ & \cellcolor{cyan!20} $\mathbf{15.1 \pm 4.3}$~\textbf{\si{\milli\meter}}  & \cellcolor{cyan!20} $\mathbf{0\%}$ & \revtwo{\cellcolor{cyan!20} $\mathbf{17.9 \pm 2.2 s}$} \\
\midrule
\multirow{4}{*}{Exp.2} 
       & w/o SF      & $20\%$   & $-23.2 \pm 10.8$~\si{\milli\meter}   & --   &  \revtwo{$35.1 \pm 2.1 s$}  \\
       & \revtwo{DCOL-based~\cite{Dai2023RAL}} & \revtwo{$90\%$} & \revtwo{$17.4 \pm 2.5$~\si{\milli\meter}}   & \revtwo{$0\%$} &  \revtwo{$38.5 \pm 3.7 s$} \\
       & Impl.-based$^{\ddagger}$~\cite{Lukas2025SemanticallySafety} & -- & --   & $51.9 \pm 3.7\%$  & \revtwo{--}\\
       & \cellcolor{cyan!20} \textbf{SDF-based (ours)}  & \cellcolor{cyan!20} $\mathbf{100\%}$  & \cellcolor{cyan!20} $\mathbf{10.6 \pm 2.1}$~\textbf{\si{\milli\meter}}  & \cellcolor{cyan!20} $\mathbf{0\%}$ & \cellcolor{cyan!20} \revtwo{$\mathbf{32.4 \pm 4.1 s}$} \\
\bottomrule
\end{tabular}
}

\vspace{1mm}
\footnotesize
\raggedright
\revtwo{
\noindent
$^{\dagger}$~Completion-time statistics are computed over successful trials only.
} \\
\revone{
$^{\ddagger}$~Evaluated offline by replaying the successful SDF-based trials.
}

}
\end{table}

\section{Conclusion}

In this work, we presented a geometry-aware safety filter 
that combines expressive SQ-based collision models with an SDF-based formulation.
By resolving the numerical issues inherent to implicit superquadric functions, the proposed approach bridges the gap between high-fidelity geometric modeling and reliable, real-time gradient-based safety filtering.
Extensive simulations and real-world experiments demonstrate that the proposed safety filter consistently achieves collision-free execution under challenging geometric conditions, sensing noise, and dynamic disturbances, while also improving task efficiency in teleoperated manipulation by reducing unnecessary corrective motions.


%%%%%%%%%%%%%%%%%%%%%%%%%%%%%%%%%%%%%%%%%%%%%%%%%%%%%%%%%%%%%%%%%%%%%%%%%%%%%%%%

%\section*{ACKNOWLEDGMENT}

%The preferred spelling of the word ÒacknowledgmentÓ in America is without an ÒeÓ after the ÒgÓ. Avoid the stilted expression, ÒOne of us (R. B. G.) thanks . . .Ó  Instead, try ÒR. B. G. thanksÓ. Put sponsor acknowledgments in the unnumbered footnote on the first page.

%%%%%%%%%%%%%%%%%%%%%%%%%%%%%%%%%%%%%%%%%%%%%%%%%%%%%%%%%%%%%%%%%%%%%%%%%%%%%%%%

\bibliographystyle{IEEEtran}

\bibliography{ref}

\end{document}